\documentclass[10pt,twocolumn,letterpaper]{article}

\usepackage{iccv}
\usepackage{times}
\usepackage{epsfig}
\usepackage{graphicx}
\usepackage{amsmath}
\usepackage{amssymb}
\usepackage{multirow}
\usepackage{array}
\usepackage{booktabs}
\usepackage{tablefootnote}
\usepackage{eso-pic}
\usepackage{setspace}
\usepackage[accsupp]{axessibility}
\usepackage[para,symbol,hang]{footmisc}

\usepackage[pagebackref=true,breaklinks=true,letterpaper=true,colorlinks,bookmarks=false]{hyperref}

\usepackage[capitalize]{cleveref}
\crefname{section}{Sec.}{Secs.}
\Crefname{section}{Section}{Sections}
\Crefname{table}{Table}{Tables}
\crefname{table}{Tab.}{Tabs.}

\iccvfinalcopy 

\def\iccvPaperID{2636} 

\ificcvfinal\fi

\begin{document}

\title{CoTDet: Affordance Knowledge Prompting for Task Driven Object Detection}

\author{
Jiajin Tang\thanks{Both authors contributed equally.} \quad 
Ge Zheng\footnotemark[1] \quad 
Jingyi Yu \quad 
Sibei Yang\thanks{Corresponding author.} \\
School of Information Science and Technology, ShanghaiTech University \\
 {\tt\small \{tangjj,zhengge,yujingyi,yangsb\}@shanghaitech.edu.cn} \vspace{-0mm} \\
\small\url{https://toneyaya.github.io/cotdet}
}

\maketitle
\ificcvfinal\thispagestyle{empty}\fi

\maketitle
\ificcvfinal\thispagestyle{empty}\fi
\begin{abstract}
Task driven object detection aims to detect object instances suitable for affording a task in an image. Its challenge lies in object categories available for the task being too diverse to be limited to a closed set of object vocabulary for traditional object detection. 
Simply mapping categories and visual features of common objects to the task cannot address the challenge.  
In this paper, we propose to explore fundamental affordances rather than object categories, i.e., common attributes that enable different objects to accomplish the same task. Moreover, we propose a novel multi-level chain-of-thought prompting (MLCoT) to extract the affordance knowledge from large language models, which contains multi-level reasoning steps from task to object examples to essential visual attributes with rationales. 
Furthermore, to fully exploit knowledge to benefit object recognition and localization, we propose a knowledge-conditional detection framework, namely CoTDet. It conditions the detector from the knowledge to generate object queries and regress boxes. 
Experimental results demonstrate that our CoTDet outperforms state-of-the-art methods consistently and significantly (+15.6 box AP and +14.8 mask AP) and can generate rationales for why objects are detected to afford the task.

\vspace{-0.2cm}
\end{abstract}
\section{Introduction}
\label{sec:intro}
\begin{figure}[t]
\centering
\includegraphics[width=0.99\columnwidth]{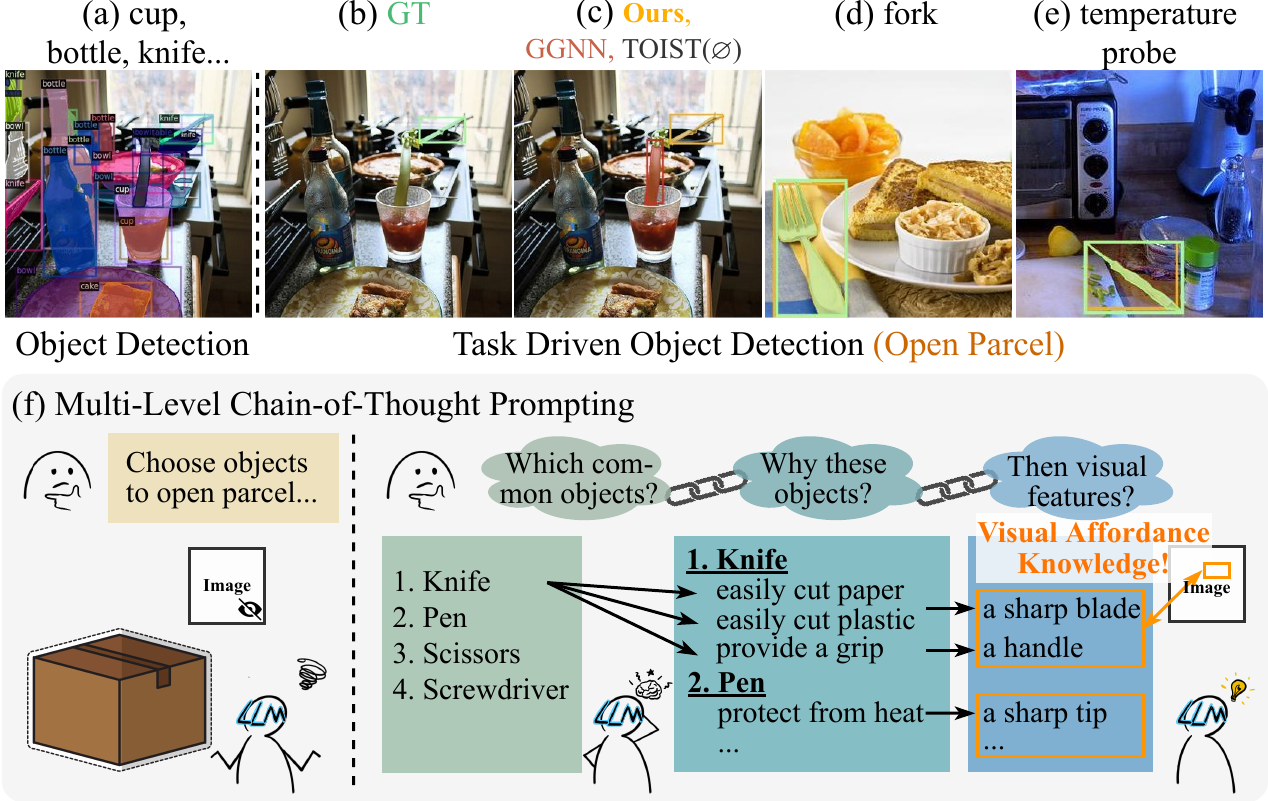}
\vspace{-0.1cm}
\caption{\textbf{An example} of (a) traditional object detection, (b)-(e) task driven object detection, and (f) our multi-level chain-of-thought (MLCoT) prompting large language models (LLMs) to generate visual affordance knowledge.}
\label{fig:intro}
\end{figure}

The traditional object detection task~\cite{he2017mask, redmon2018yolov3, carion2020end, zhu2020deformable}, as shown in Figure~\ref{fig:intro}\textcolor{red}{a}, aims to detect object instances of given categories in an image, so objects of categories such as the cup, bottle, cake, and knife are detected. In contrast, detection tasks in real-world applications~\cite{autonomous2020, ahn2022can}, such as intelligent service robots, usually appear in the form of tasks rather than object categories~\cite{ren2023leveraging}. 
For example, when an intelligent robot is asked to complete the task of \textit{``opening parcels"}, the robot needs to autonomously locate the tool shown in Figure~\ref{fig:intro}\textcolor{red}{b}, \ie, a knife. 
So the core of this type of task is to detect the object instances in the image that are most suitable for the task~\cite{sawatzky2019ggnn, li2022toist}. However, the challenges for task driven object detection are multi-fold. Previous methods that directly learn the mapping between objects and tasks from objects' visual context features or categories cannot achieve satisfactory results. As shown in Figure~\ref{fig:intro}\textcolor{red}{c}, the context-based approach GGNN~\cite{sawatzky2019ggnn} wrongly chooses \textit{vegetable stem} for afford task of \textit{``opening parcels"} because it learns that visually slender objects can open parcels. 
Similarly, the category-based approach TOIST~\cite{li2022toist} considers that no object in the image can perform the \textit{``opening parcels"} since it could not even recognize an instance of the knife in the image. In contrast, people will intelligently and naturally choose to use the knife to open parcels in the scene of Figure~\ref{fig:intro}\textcolor{red}{b}.

Recently, Large Language Models (LLMs) like GPT-3~\cite{brown2020language} and ChatGPT~\cite{ouyang2022training} have demonstrated impressive capabilities in encoding general world knowledge from a vast amount of text data~\cite{petroni2019language, roberts2020much, wei2022chain}. 
A naive approach is to prompt LLMs directly to return ``what objects should we use to \textit{open parcels}" and leverage the returned object categories to simplify task driven object detection to the traditional one.
However, we observe that LLMs usually only return a few categories of commonly used objects, such as the knife, pen, paper cutter, and scissors shown in Figure~\ref{fig:intro}\textcolor{red}{f}. 
According to these categories, although the knife in Figure~\ref{fig:intro}\textcolor{red}{b} can be identified as the target object, the detection system will miss other objects that also can be used to open parcels, such as the fork in Figure~\ref{fig:intro}\textcolor{red}{d} and the temperature probe next to the microwave oven in Figure~\ref{fig:intro}\textcolor{red}{e}. 
In turn, we ask why people can easily lock the fork and temperature probe in Figure~\ref{fig:intro}\textcolor{red}{d} and~\ref{fig:intro}\textcolor{red}{e} as the target objects? We argue that the reason is that people are not restricted to using specific categories of objects to accomplish a task but instead select objects based on the commonsense knowledge that objects with \textit{``a handle and sharp blade"} \textcolor{black}{or \textit{``a handle and sharp tip"}} can \textit{``open parcels"}.

In this paper, we propose to explicitly acquire visual affordance knowledge of the task (\ie, common visual attributes that enable different objects to afford the task) and utilize the knowledge to bridge the task and objects.  
Figure~\ref{fig:intro}\textcolor{red}{f} shows \textcolor{black}{two sets of} visual affordance knowledge (marked inside the yellow box) for \textcolor{black}{\textit{opening parcels}}. However, it is not trivial to acquire such task-specific visual affordance knowledge. 

Furthermore, we propose a novel multi-level chain-of-thought prompting (MLCoT) to elicit visual affordance reasoning from LLMs. At the first level (object level), we prompt LLMs to return common objects by the above-mentioned approach. Unlike before, which treats the returned object categories as target categories, we instead treat this query progress as brainstorming to obtain representative object examples. 
At the second level (affordance level), we generate rationales from LLMs for why object examples can afford the task and cooperate rationales to facilitate LLMs to reason and summarize the visual affordances beyond object examples. 
As shown in Figure~\ref{fig:intro}\textcolor{red}{f}, the rationale and visual affordances that enable the knife to open parcels are \textit{``easily cut through paper and plastic..."} and \textit{``a sharp blade and handle"}, respectively. 
Our MLCoT can capture the essence of visual affordances behind these object examples without being limited to object categories. Thus we can successfully detect the fork and temperature probe in Figure~\ref{fig:intro}\textcolor{red}{d} and~\ref{fig:intro}\textcolor{red}{e} as they meet the visual affordances required by the task. 

Moreover, we claim that visual affordance knowledge not only helps recognize and identify objects suitable for the task but also helps localize objects more precisely because visual attributes such as color and shape are useful in object localization. 
Therefore, unlike some methods~\cite{singh2018dock,shen2022k,wu2016ask,narasimhan2018out,gao2022transform} to take knowledge as complementary to the image's visual features, we condition the detector on the visual affordance knowledge to perform knowledge-conditional object detection. 
Specifically, we follow~\cite{li2022toist} to use an end-to-end query-based detection framework~\cite{carion2020end,zhu2020deformable}. But instead of randomly initializing queries, we generate knowledge-aware queries based on image features and visual affordance knowledge.
In addition to generating queries, we use visual affordance knowledge to guide the bounding box regression explicitly, inspired by the denoising training~\cite{li2022dn}. Unlike \cite{li2022dn} introduces denoising for accelerating training, our knowledge-conditional denoising training aims to teach the decoder how to utilize visual knowledge to regress the boxes for queries.

Finally, we propose the CoTDet network, which acquires visual affordance knowledge from LLMs via the proposed MLCoT and performs knowledge-conditional object detection to effectively utilize the knowledge. 
Moreover, our CoTDet can easily be extended to task driven instance segmentation by employing a segmentation head~\cite{cheng2021per,kamath2021mdetr}. 

In summary, our main contributions are:
\begin{itemize}
\setlength{\itemsep}{0pt}
\setlength{\parsep}{0pt}
\setlength{\parskip}{0pt}

\item We are the first to propose to explicitly acquire visual affordance knowledge and utilize the knowledge to bridge the task and object instances. 
 \item We propose a novel multi-level CoT prompting (MLCoT) to make abstract affordance knowledge concreted, which leverages LLMs to generate and summarize intermediate reasoning steps from object examples to essential visual attributes with rationales. 
        \item We claim that visual affordance knowledge can benefit both object recognition and localization and propose a knowledge-conditional detection framework to condition the detector to generate object queries and guide box regression through denoising training. 
        \item Our CoTDet not only consistently outperforms state-of-the-art methods ($+15.6$ $\text{AP50}_{\text{box}}$ and $+14.8$ $\text{AP50}_{\text{mask}}$) by a large margin but also can generate rationales for why objects are detected to afford tasks.
\end{itemize}
\begin{figure*}[t]
\centering
\includegraphics[width=0.98\textwidth]{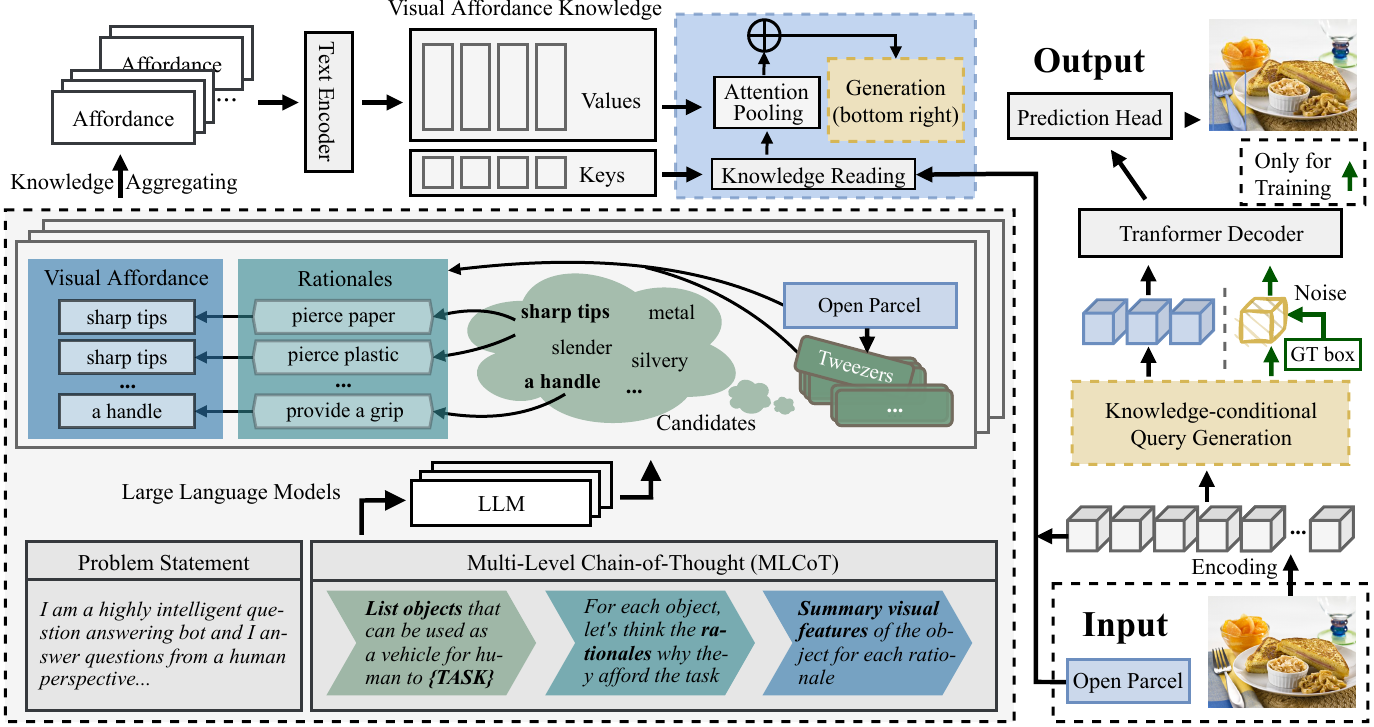}
\caption{\textbf{Overall framework of the proposed CoTDet with multi-level chain-of-thought prompting (MLCoT).} We first generate visual affordance knowledge from LLMs with the proposed novel MLCoT. Next, we perform knowledge-conditional object detection by utilizing the knowledge to generate object queries for the scene as well as guide object localization through denoising training.
}
\vspace{-0.3cm}
\label{fig:framework}
\end{figure*}

\section{Related Work}
\noindent \textbf{Task Driven Object Detection/Instance Segmentation} aims to detect or segment out the most suitable objects in an image to afford a given task, such as \textcolor{black}{\textit{opening parcel} or \textit{getting lemon out of tea}}. Different from traditional object detection or instance segmentation~\cite{redmon2018yolov3,girshick2015fast,zhu2020deformable,carion2020end,he2017mask,cheng2021per}, it requires modeling the preference for selecting objects based on a comprehensive understanding of the specific task and the image scene. 
Although referring expression grounding~\cite{deng2021transvg, kamath2021mdetr, rohrbach2016grounding,chengshi,yang2020propagating,yang2019cross,yang2020relationship} and segmentation~\cite{yang2022lavt,yang2021busnet,wang2022cris,hu2016segmentation,yu2018mattnet,lin2021structured,yang2019dynamic,cgformer} similarly locate the referent according to a natural language description, they rely on the object attributes and relations in the scene for localization without considering the prior knowledge needed to afford the given task. 
These factors result in the challenge of task driven object detection which requires complex and joint knowledge reasoning on the requirements for one specific task and objects' functional attributes beyond visual recognition and scene understanding.

The pioneering work~\cite{sawatzky2019ggnn} adopts a two-stage framework that first detects objects and then compares among objects to select suitable objects via graph neural networks~\cite{scarselli2008graph}. 
The following work TOIST~\cite{li2022toist} distills target object names to pronouns such as something by conditioning on the language description of the task. However, these works lack to model explicit requirements of tasks and objects' affordances to tasks, which limits their performance and generalization capabilities.

\noindent \textbf{Knowledge Acquisition in Vision-Language Tasks.} Integrating external knowledge into computer vision tasks~\cite{singh2018dock, qi2019ke, kato2018compositional, shen2022k} and vision-language tasks~\cite{zareian2020bridging, wang2017fvqa, marino2019ok, vo2022noc} has been found beneficial. 
Previous methods~\cite{wu2016ask, narasimhan2018out, marino2021krisp, qi2019ke,kato2018compositional, yu2022probabilistic} are interested in acquiring structured knowledge (e.g., ConceptNet~\cite{liu2004conceptnet}), which usually includes commonsense concepts and relations and is presented in fixed data structures such as the graph or triple. 
Recently, large language models~\cite{brown2020language,raffel2020exploring,devlin2018bert} have been demonstrated to learn open-world commonsense knowledge from the large-scale corpus~\cite{petroni2019language, roberts2020much}. 
Some works~\cite{yang2022empirical, gao2022transform} utilize language models to encode the representations of inputs or directly generate answers conditioning on visual inputs, leveraging the latent knowledge in language models. 
Unlike previous works, we prompt language model GPT-3~\cite{floridi2020gpt} to obtain external knowledge explicitly with the chain of thought (CoT)~\cite{wei2022chain, zhang2023multimodal, lu2023dynamic} for better interpretability. 
To the best of our knowledge, we are the first to explore CoT to acquire visual commonsense knowledge in text form, leveraging the reasoning ability of the language model to filter effective visual functional attributes to afford tasks. 

\noindent \textbf{External Knowledge Utilization.}
Incorporating knowledge reasoning has attracted growing interest in computer vision~\cite{shen2022k,mao2022doubly,pratt2022does,qi2019ke,zareian2020bridging,gu2019scene} and vision-language~\cite{vo2022noc} fields such as object detection~\cite{singh2018dock,shen2022k}, visual relationship detection~\cite{kato2018compositional,yu2022probabilistic} and visual question answering~\cite{wu2016ask,narasimhan2018out,marino2021krisp,gao2022transform}. 
For tasks that focus on capturing relations among objects, such as scene graph generation and visual relationship detection, extracting knowledge of the interactions between general concepts becomes natural~\cite{zareian2020bridging,gu2019scene,kato2018compositional,yu2022probabilistic}. Other tasks like object detection and image classification rely on category-related knowledge that is retrieved in the knowledge base as definitions or attributes of general concepts~\cite{singh2018dock,shen2022k,wu2016ask,gao2022transform,mao2022doubly,pratt2022does}.
For knowledge utilization, they mainly directly take external knowledge as an expansion to visual content or explicitly constrain the consistency between knowledge and visual content. Different from existing works, we leverage attribute-level commonsense knowledge about requirements for completing tasks and take external knowledge of tasks as the condition to condition the detector for task driven object detection.

\section{Method}
The framework of our proposed CoTDet is shown in Figure~\ref{fig:framework}. First, we introduce the problem definition and image and text encoders in Section~\ref{sec:m0}. Second, we acquire visual affordance knowledge from LLMs by leveraging the multi-level chain-of-thought prompting and aggregation (Section~\ref{sec:m1}). Next, we present the knowledge-conditional decoder that conditions acquired knowledge to detect and segment suitable objects (Section~\ref{sec:m2}). Finally, we introduce the loss functions in Section~\ref{sec:loss}.

\subsection{Problem Definition and Encoder}
\label{sec:m0}
\noindent \textbf{Problem Definition.} Given a task in text form $S$ (\eg, \textit{``open parcel"}) and an image $I$, task driven object detection require detecting a set of objects most preferred to afford the task. Note that the target objects indicated by the task are not fixed in quantity and category, which may vary with changes in the image scene. In contrast, traditional object detection~\cite{girshick2015fast, carion2020end, zhu2020deformable} detects objects of fixed categories while referring image grounding~\cite{plummer2017phrase, mao2016generation, kazemzadeh2014referitgame, deng2021transvg} localizes unambiguous objects.

\noindent \textbf{Encoders.}
For the task $S$, we leverage the linguistic information from two perspectives. First, the text $S$ is preserved as input for extracting knowledge from LLMs. Besides, we employ RoBERTa~\cite{liu2019roberta} as the text encoder to obtain the text feature $t_s$ for the task $S$, which will be used to query the task-relevant visual content in the image. 
For the image $I$, we adopt the ResNet-101~\cite{he2016deep} as the visual backbone to extract multi-scale visual feature maps and flatten the maps along the spatial dimension to features $V$. 

\subsection{Visual Affordance Knowledge from LLMs}
\label{sec:m1}
To detect objects to afford a particular task, we naturally consider the task's requirements first and subsequently localize suitable objects that meet the requirements. Nevertheless, the task requirements are abstract and can not directly correspond to the visual content in the image for localizing the objects. Motivated by this, we propose to explicitly extract common visual attributes (\ie, visual affordances) that make different objects can afford the task and use visual affordances to bridge task requirements and object instances in the image. Furthermore, we generate visual affordances from LLMs because LLMs generally contain world knowledge learned from a vast amount of text data.

Specifically, we first design a novel multi-level chain-of-thought prompting to leverage LLMs to generate visual affordances (see Section~\ref{sec:mlcot}) and then encode and aggregate them automatically to be utilized for detection (see Section~\ref{sec:kea}). 

\subsubsection{Multi-Level Chain-of-Thought Prompting}
\label{sec:mlcot}
Our multi-level chain-of-thought prompting (MLCoT) leverages LLMs to generate and summarize intermediate reasoning steps from object examples
to essential visual attributes with rationales. MLCoT first brainstorms the object examples to afford the task and then considers rationales why the examples can afford the task and summarizes corresponding visual attributes for rationales.

\noindent\textbf{Object-level Prompting as Brainstorming.} At the first level, we prompt LLMs to generate daily object examples that afford the input task $S$. Specifically, we design the following text prompt: 

\vspace{0.4em}
\noindent\textit{
Prompt: What common objects in daily life can be used as a vehicle for the human to \textcolor{black}{[task]}? Please list the twenty most suitable objects.}

\noindent\textit{
Output: knife, pen, paper cutter, scissors, screwdriver, ...}

\vspace{0.4em}
\noindent Where [task] is filled with the task text $S$. We denote the number of objects returned from LLMs as $N_o$. 
For simplicity, we present the most critical parts of the prompts. For complete prompts, please refer to \textcolor{black}{Appendix}. 
One straightforward idea is to perform traditional category-specific object detection with respect to the categories of object examples. However, it is not feasible due to the following observations: (1) object examples are overly limited to partial object categories, resulting in the gap between object categories and the actual task requirements. \textcolor{black}{For instance, the fork in Figure \ref{fig:framework} is not among the objects returned by LLMs.} (2) a few noisy unsuitable objects may be output. Although the noisy objects are few, relying entirely on the object examples is risky. For example, for the task of extinguishing fire, LLMs return the fire axe, a common firefighting tool, but it cannot be directly used to extinguish the fire. 

\noindent\textbf{Affordance-level Prompting with Rationales.} 
To address the above challenges and capture the essential visual affordances implied in representative object examples, we propose to generate rationales for why these objects can afford the task and summarize visual affordances from rationales. With the object examples, we prompt LLMs to generate rationales as follows:

\vspace{0.4em}
\noindent\textit{
Prompt: For each object, let's think about the rationales for why they afford the task from the perspective of visual features.}

\noindent\textit{
Output: Knife: They have a sharp blade which can easily cut through paper and plastic; They have a handle which provides a good grip for the user. 
Pen: ... }

\vspace{0.4em}

\noindent By the above prompting, we get a set of rationales for each object. Next, we further prompt LLMs to summarize visual-related rationales to form visual affordances as follows: 

\vspace{0.4em}

\noindent\textit{
Prompt: Summary corresponding visual features of the object for each rationale.}

\noindent\textit{
Output: \{A sharp blade and a handle. \}, \{...\}, ...}

\vspace{0.4em}
\noindent \textcolor{black}{Finally, we obtain $N_o$ sets of visual affordances, where each set contains several visual attributes relevant to why objects can afford the task $S$. And we define each set of visual affordance knowledge as a knowledge unit.} Note that although each knowledge unit is derived from the rationales of an object example, the affordance knowledge in that unit is not limited to that object example or its related categories. \textcolor{black}{For example, the first visual affordance unit comprises \textit{``a sharp blade and a handle"}, which correspond to the returned object \textit{``knife"}. Notably, this visual affordance unit also applies to \textit{``box cutter"} and \textit{``paper cutter"}.}

\begin{table*}[t]
\small
\centering

\setlength{\tabcolsep}{0.98mm}{\begin{tabular}{ccccc}
\toprule[1pt]
\multicolumn{1}{l}{task1: \textit{step on}}&\multicolumn{1}{l}{task2: \textit{sit comfortably}}&\multicolumn{1}{l}{task3: \textit{place flowers}}&\multicolumn{1}{l}{task4: \textit{get potatoes out of fire}}&\multicolumn{1}{l}{task5: \textit{water plant}}\\
\multicolumn{1}{l}{task6: \textit{get lemon out of tea}}&\multicolumn{1}{l}{task7: \textit{dig hole}}&\multicolumn{1}{l}{task8: \textit{open bottle of beer}}&\multicolumn{1}{l}{task9: \textit{open parcel}}& \multicolumn{1}{l}{task10: \textit{serve wine}}\\
\multicolumn{1}{l}{task11: \textit{pour sugar}}& \multicolumn{1}{l}{task12: \textit{smear butter}}& \multicolumn{1}{l}{task13: \textit{extinguish fire}}& \multicolumn{1}{l}{task14: \textit{pound carpet}}&\\
\noalign{\smallskip}
\end{tabular}}

\setlength{\tabcolsep}{0.75mm}{\begin{tabular}{l|cccccccccccccc|c}
\toprule[1pt]
\multirow{2}{*}{Method}  & \multicolumn{14}{c|}{Task(AP$_{\text{box}}$@0.5)} & \multirow{2}{*}{Mean}  \\
\cline{2-15}
\rule{0pt}{10pt} 
 & \phantom{1}task1 & \phantom{1}task2 & \phantom{1}task3 & \phantom{1}task4 & \phantom{1}task5 & \phantom{1}task6 & \phantom{1}task7 & \phantom{1}task8 & \phantom{1}task9 & task10 & task11 & task12 & task13 & task14 &\\
\midrule
GGNN~\cite{sawatzky2019ggnn}&36.6&29.8&40.5&37.6&41.0&17.2&43.6&17.9&21.0&40.6&22.3&28.4&39.1&40.7&32.6\\
TOIST~\cite{li2022toist}&44.0& 39.5 & 46.7 & 43.1 & 53.6 & 23.5 & 52.8 & 21.3 & 23.0 & 46.3 & 33.1 & 41.7 & 48.1 & 52.9 & 41.3 \\ 
TOIST$^{\dagger}$~\cite{li2022toist}&45.8& 40.0 & 49.4 & 49.6 & 53.4 & 26.9 & 58.3 & 22.6 & 32.5 & 50.0 & 35.5 & 43.7 & 52.8 & 56.2 & 44.1 \\ 
Ours  &\textbf{58.9}&\textbf{55.0}&\textbf{51.2}&\textbf{68.5}&\textbf{60.5}&\textbf{47.7}&\textbf{76.9}&\textbf{40.7}&\textbf{47.4}&\textbf{66.5}&\textbf{41.9}&\textbf{48.3}&\textbf{61.7}&\textbf{71.4}&\textbf{56.9} \\ 
\bottomrule[1pt]
\end{tabular}}
\medskip
\caption{\textbf{Comparison with state-of-the-art models for task driven object detection on COCO-Tasks dataset.} $^{\dagger}$ means the model is with noun-pronoun distillation.}
\label{tab:det}
\vspace{-0.2cm}
\end{table*}
\subsubsection{Knowledge Encoding and Aggregating}
\label{sec:kea}
We further extract a refined knowledge base by filtering out a few knowledge units corresponding to the unsuitable object examples mentioned in Section~\ref{sec:mlcot}. 
For each knowledge unit, we concatenate textual descriptions of its visual affordances into a textual sequence and then utilize RoBERTa~\cite{liu2019roberta} to obtain the sentence feature. To filter out unsuitable units, we compute the cosine similarity between each pair of knowledge units and exclude outlier units if their maximum similarity to other units falls below a predetermined threshold. Additionally, for each selected unit, we extract its word representations via RoBERTa. In summary, we aggregate $N$ visual affordance knowledge units, denoted as $\mathcal{K} = \{p^k_j,p^v_j\}_{j=1}^N$, where $p^k_j$ and $p^v_j$ are the sentence feature and word features of $j$-th unit, respectively. 

\subsection{Knowledge-conditioned Decoder}
\label{sec:m2}
We base on the detection architecture of Deformable-DETR~\cite{zhu2020deformable}, a DETR-like detector~\cite{carion2020end, li2022dn, cheng2021per}, which uses object queries to capture object-level information for detection (Section~\ref{sec:detr}). Unlike randomly initializing the object queries, we leverage visual affordance knowledge to generate the object queries (Section~\ref{sec:query}) and guide the bounding box regression with denoising training (Section~\ref{sec:decoding}). 

\subsubsection{Introduction to Deformable-DETR}
\label{sec:detr}

Deformable-DETR contains a Transformer encoder and a Transformer decoder. The encoder inputs visual features $V$ and outputs the refined visual features $F=\{f_1, f_2, ..., f_i\}$ via multi-scale deformable attention. The decoder randomly initializes queries $Q=\{q_1, q_2, ..., q_k\}$ and predicts a reference point $p_k$ for each object query $q_k$, and these reference points $L=\{l_1, l_2, ..., l_k\}$ serve as the initial guess of the box centres. Next, the decoder searches for objects $O$ for these queries $Q$ with reference points $L$ via multi-scale deformable cross-attention and self-attention, which is formulated as follows, 
\begin{equation}
    O = \mathrm{Deformable}([Q, L], F),
\end{equation}
where $\mathrm{Deformable}(\cdot, \cdot)$ denotes the Transformer decoder of Deformable-DETR.

\subsubsection{Knowledge-conditional Query Generation}
\label{sec:query}
Instead of randomly initializing the queries, we generate the queries and their reference points based on the visual content of the image, the task, and the visual affordance knowledge. Specifically, we utilize visual affordance knowledge to select visual features and combine them with the knowledge to generate queries, and then the spatial information in these visual features naturally becomes the spatial priors of reference points.

Given visual features $F=\{f_1, f_2, ..., f_i\}$, we first fuse each feature $f_i$ with the task's text feature $t_s$, and then calculate its relevance to the task's visual affordance knowledge $\mathcal{K} = \{(p^k_j,p^v_j)\}_{j=1}^N$. Since each knowledge unit $(p^k_j,p^v_j)$ in the knowledge base $\mathcal{K}$ is a set of affordances that meet the task requirements, we use the fused feature's largest similarity to each knowledge unit $(p^k_j,p^v_j)$ in the knowledge base $\mathcal{K}$ as the feature's relevance score. The calculation is formulated as follows,

\begin{equation}
\label{eq:read}
    \begin{aligned}
    &s_{i,j} = \mathrm{cos}(\mathrm{fc}(f_i) + \mathrm{fc}(t_s), p_j^k), \\
    &r_i = \mathrm{max}_j(s_{i,j}), d_i = \mathrm{argmax}_j(s_{i,j}),
    \end{aligned}
\end{equation}
where $\mathrm{cos}(\cdot, \cdot)$ computes the cosine similarity, $\mathrm{fc}(\cdot)$ represents the fully connected layer, and $s_{i,j}$ is the similarity between $i$-th visual feature and $j$-th knowledge unit. Then, $r_i$ and $d_i$ mean the $i$-th visual feature's relevance score and index of the corresponding knowledge unit, respectively. 

Next, we select the visual features with the top-$k$ largest relevance scores $\{r_i\}$ to incorporate their corresponding knowledge $\{(p^k_{d_i}, p^v_{d_i})\}$ to generate queries $Q^\text{kn}$ as follows,
\begin{equation}
\label{eq:query}
    Q^{\text{kn}} = \mathrm{topk}_{r_i}\{f_i + \mathrm{AttentionPool}(f_i, p^v_{d_i})\},
\end{equation}
where $\mathrm{topk}_{r_i}$ means to select the corresponding features with the top-$k$ largest relevance scores $r_i$. The attention pooling layer~\cite{vaswani2017attention} $\mathrm{AttentionPool}(f_i, p^v_{d_i})$ returns the weighted features on $p^v_{d_i}$ based on their similarities to the $f_i$. Note that, for each knowledge unit $(p^k_j,p^v_j)$, we use its global sentence feature $p^k_j$ to compute its overall similarity to each visual feature in Eq.~\ref{eq:read} while adopting word-level features $p^v_j$ to better enhance the query's fine-grained representations in Eq.~\ref{eq:query}. 
Similar to Deformable-DETR, we further predict the reference points $L^{\text{kn}}$ from the queries $Q^{\text{kn}}$.
In addition, to facilitate the learning of Top-$k$ selection, \textcolor{black}{the selected queries $Q^{\text{kn}}$ are directly fed into the prediction heads and supervised during training using the same training loss in Section~\ref{sec:loss}. }

\subsubsection{Knowledge-conditional Decoding}
\label{sec:decoding}
With queries $Q^{\text{kn}}$, reference points $L^{\text{kn}}$, and the refined visual features $F$, we apply the Deformable decoder to search objects $O^{\text{kn}}$ as follows,
\begin{equation}
\label{eq:normal}
    O^{\text{kn}}=\mathrm{Deformable}([Q^{\text{kn}},L^{\text{kn}}], F).
\end{equation}
In addition to utilizing visual affordance knowledge for query generation and providing the decoder with prior knowledge, we further improve the knowledge utilization by designing a knowledge-based denoising training~\cite{li2022dn}. As the visual affordance knowledge indicates the target objects' visual attributes, such as shape and size, the knowledge-base denoising guides the decoder in learning how to use this kind of visual knowledge to regress the targets' boxes. 

Specifically, during the training stage, we first randomly add noise to ground-truth boxes $O^{\text{gt}}=\{o^{\text{gt}}_m\}_{m=1}^M$ 
to construct the noised objects following DN-DETR~\cite{li2022dn} and then extract noised boxes' visual features and centers as the noised queries $F^\text{noise}=\{f^{\text{noise}}_m\}_{m=1}^M$. Notice that the previous denoising training method~\cite{li2022dn} adds noise to both boxes and categories labels to capture label-box relations better. But we only add noise to boxes because we aim to utilize the knowledge without noise to help denoise boxes.
Therefore, we extract the knowledge unit $(p^k_{d_m}, p^v_{d_m})$ for each ground-truth box $o^{\text{gt}}_m$ through Eq.~\ref{eq:read}. Finally, the knowledge units $\{(p^k_{d_m}, p^v_{d_m})\}_{m=1}^M$ guide the decoder to regress the ground-truth boxes $O^{\text{gt}}$ from the noised queries $F^\text{noise}$, which is formulated as follows, 
\begin{equation}
\begin{aligned}
    \label{eq:kn}
    P^{\text{kn}} &= \{\mathrm{AttentionPool}(f_m^{\text{noise}}, p^v_{d_m})\}_{m=1}^M \\
    O^{\text{denoise}}&=\mathrm{Deformable}([F^{\text{noise}} + P^{\text{kn}}, L^\text{noise}], F,) \\
\end{aligned}
\end{equation}
where $P^{\text{kn}}$ is the visual affordance knowledge of noised queries, and the $\mathrm{Deformable}(\cdot, \cdot)$ in Eq.~\ref{eq:normal} and Eq.~\ref{eq:kn} shares the same parameters. And the denoising is only considered in the training stage. 

\subsection{Loss Functions}
\label{sec:loss}
Following DETR~\cite{carion2020end}, we use bipartite matching to find the unique predictions for the ground-truth objects and adopt the same bounding box regression loss $\mathcal{L}_{box}$ consisting of L1 loss and GIoU~\cite{rezatofighi2019generalized} loss. Moreover, we use the binary cross entropy loss as the classification loss $\mathcal{L}_{cl}$. The overall loss is represented as:
\begin{equation}
\mathcal{L}_{cost} =  \lambda_{cl}\mathcal{L}_{cl} +\lambda_{box}\mathcal{L}_{box}, 
\end{equation}
where $\lambda_{cl}$ and $\lambda_{box}$ are the hyperparameters of the weighted loss. Our method can be easily extended to instance segmentation by adding a segmentation head~\cite{cheng2021per} and replacing the box regression loss with the Dice loss $\mathcal{L}_{mask}$.
\begin{table*}[t]
\small
\centering
\setlength{\tabcolsep}{0.75mm}{\begin{tabular}{l|cccccccccccccc|c}
\toprule[1pt]
\multirow{2}{*}{Method} & \multicolumn{14}{c|}{Task(AP$_{\text{mask}}$@0.5)} & \multirow{2}{*}{Mean}  \\
\cline{2-15}
\rule{0pt}{10pt} 
 & \phantom{1}task1 & \phantom{1}task2 & \phantom{1}task3 & \phantom{1}task4 & \phantom{1}task5 & \phantom{1}task6 & \phantom{1}task7 & \phantom{1}task8 & \phantom{1}task9 & task10 & task11 & task12 & task13 & task14 &\\
\midrule
GGNN~\cite{sawatzky2019ggnn}&31.8&28.6&45.4&33.7&46.8&16.6&37.8&15.1&15.0&49.9&24.9&18.9&49.8&39.7&32.4\\
TOIST~\cite{li2022toist}&37.0&34.4&44.7&34.2&51.3&18.6&40.5&17.1&23.4&43.8&29.3&39.9&46.6&42.4&35.2 \\  
TOIST$^{\dagger}$~\cite{li2022toist} &40.8&36.5&48.9&37.8&43.4&22.1&44.4&20.3&26.9&48.1&31.8&34.8&51.5&46.3&38.8 \\
Ours  &\textbf{55.0}&\textbf{51.6}&\textbf{51.2}&\textbf{57.7}&\textbf{60.1}&\textbf{43.1}&\textbf{65.9}&\textbf{40.4}&\textbf{45.4}&\textbf{64.8}&\textbf{40.4}&\textbf{48.7}&\textbf{61.7}&\textbf{64.4}&\textbf{53.6}\\
\bottomrule[1pt]
\end{tabular}}
\medskip
\caption{\textbf{Comparison with state-of-the-art models for task driven instance segmentation on COCO-Tasks dataset.} $^{\dagger}$ means the model is with noun-pronoun distillation.}
\label{tab:seg}
\vspace{-0.2cm}
\end{table*}
\begin{table}[t]
    \begin{center}
    \small
    \setlength{\tabcolsep}{0.6mm}{\begin{tabular}{cccccc}
    \toprule
    \textbf{Ablation}& \phantom{1}task2\phantom{1}& \phantom{1}task6\phantom{1}& \phantom{1}task9\phantom{1}& \textbf{mAP}$_{\text{box}}$ & \textbf{mAP}$_{\text{mask}}$  \\
    \midrule
    \textit{``objects''}&25.4& 16.5& 21.0&31.9&31.3 \\ 
    \textit{``visual''}  &50.4&30.3&38.3&48.1&44.7\\
    w/o rationales        &52.0&40.7&41.2&52.4&49.0\\
    MLCoT    &55.0&47.7&47.5&56.9&53.6\\
    MLCoT(ChatGPT)      &50.6&48.1&50.3&57.0&54.0\\
    \midrule
    Def+GGNN~\cite{sawatzky2019ggnn}   &38.6&24.7&23.4&38.8&35.8\\ 
    Def+TOIST~\cite{li2022toist}   &43.4&21.0&29.0&40.3&37.6\\
    \midrule                          
    Init w/ MLCoT       &42.2&35.9&35.6&48.7&46.4\\ 
    Fuse w/ MLCoT       &44.0&42.3&41.2&50.6&47.7\\ 
    Select w/ MLCoT     &50.0&47.2&43.7&55.3&51.7\\ 
    Full Decoder        &55.0&47.7&47.5&56.9&53.6\\
    \bottomrule
    \end{tabular}}
    \end{center}
    \caption{\textbf{Ablation study} about knowledge acquisition, detection framework, and knowledge utilization of our CoTDet.}
    \label{tab:ablations}
    \vspace{-0.4cm}
\end{table}

\section{Experiment}
\subsection{Dataset and Implementation Details}
\noindent \textbf{Dataset.} We conduct experiments on the COCO-Tasks dataset~\cite{sawatzky2019ggnn}, which comprises 14 different tasks (see Table~\ref{tab:det}). This dataset is derived from the COCO dataset~\cite{lin2014coco}, but with customized annotations for task driven object detection. Each task contains 3600 training and 900 testing images. Besides, we follow~\cite{li2022toist} to incorporate mask annotations to the original COCO-Tasks dataset for the instance segmentation benchmark.

\noindent \textbf{Implementation Details.} Following previous works~\cite{sawatzky2019ggnn,li2022toist}, we use ResNet-101~\cite{he2016deep} as the image encoder and RoBERTa~\cite{liu2019roberta} as the text encoder. The model is pre-trained on the COCO dataset but images already part of COCO-Tasks are removed. We train the model for 4000 iterations with the initial learning rate 1e-4 and use AdamW~\cite{loshchilov2017adamw} as the optimizer. The hyperparameters $\lambda_{cl}$ and $\lambda_{box}$ are 4 and 5. Following~\cite{li2022toist}, we evaluate the segmentation and detection performance of each task using AP$_{\text{mask}}$@0.5 and AP$_{\text{box}}$@0.5, respectively. And we denote their means across all tasks as mAP$_{\text{mask}}$ and mAP$_{\text{box}}$. 
Unless otherwise specified, we leverage the GPT-3~\cite{brown2020language} to extract visual affordance knowledge due to its capability to generate rationales~\cite{wei2022chain}.

\subsection{Comparison with State-of-the-Art Methods}
Table~\ref{tab:det} and Table~\ref{tab:seg} show the comparison of our CoTDet with state-of-the-art models (SOTAs) on detection and segmentation benchmarks. Our model consistently outperforms the SOTAs~\cite{sawatzky2019ggnn,li2022toist} on all benchmarks and tasks.

\noindent \textbf{Comparison with SOTAs.}
Compared to TOIST~\cite{li2022toist}, our CoTDet achieves significant performance improvement (15.6\% mAP$_{\text{box}}$ and 14.8\% mAP$_{\text{mask}}$), which demonstrates the effectiveness of our task-relevant knowledge acquisition and utilization. Compared to the two-stage method GGNN~\cite{sawatzky2019ggnn}, we achieve 24.3\% mAP$_{\text{box}}$ and 21.2\% mAP$_{\text{mask}}$ performance gain, which demonstrates the importance of leveraging the visual affordance knowledge rather than purely visual context information.

\noindent \textbf{Comparison on Sub-tasks.}
The following comparisons on sub-tasks further demonstrate that the affordance-level knowledge is capable of bridging tasks and objects. 
Our CoTDet significantly improves the detection and segmentation performance on task4 (\textit{get potatoes out of fire}), task6 (\textit{get lemon out of tea}) and task7 (\textit{dig hole}), achieving approximately 20\% mAP improvement on both benchmarks. 
These tasks face the common challenge of the wide variety of targets' categories and visual appearances, which is hardly dealt with by methods like~\cite{li2022toist, sawatzky2019ggnn} that merely learn the mapping between tasks and objects' categories and visual features. In contrast, our method explicitly acquires the visual affordance knowledge of tasks to detect rare objects and avoid overfitting to common objects, outperforming significantly in these tasks. 
In addition, for those less challenging tasks with a few ground-truth object categories, we still achieve approximately 8\% mAP improvement, demonstrating the effectiveness of conditioning on visual affordances to object localization. 

\begin{figure*}[t]
\centering
\includegraphics[width=0.98\textwidth]{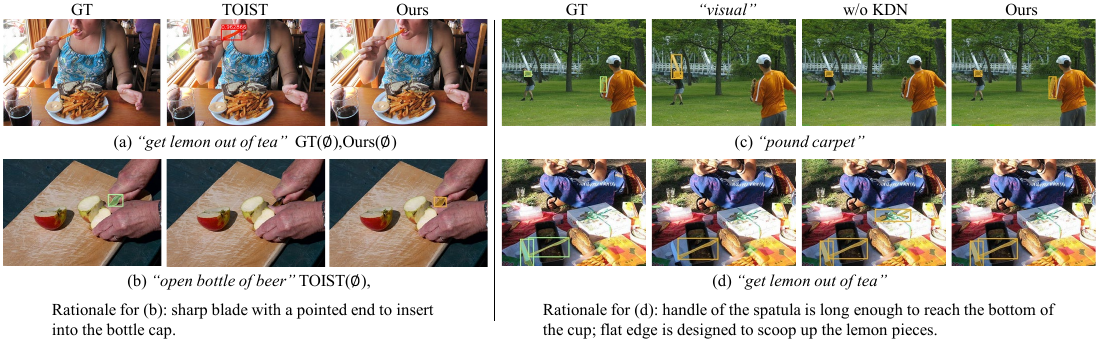}
\vspace{-0.1cm}
\caption{\textbf{Visualization for prediction results} of our CoTDet, its variants, and the existing best-performing TOIST~\cite{li2022toist}.}
\vspace{-0.3cm}
\label{fig:vz}
\end{figure*}
\subsection{Ablation Study}
We evaluate seven variants of our CoTDet and two SOTAs with the same backbones as ours to validate the effectiveness of the proposed knowledge acquisition and utilization. 
The results are shown in Table~\ref{tab:ablations}. In addition to mAP, we report AP$_{\text{box}}$@0.5 on relatively easy task2 (\textit{sit comfortably}) and challenging task6 (\textit{get lemon out of tea}) and task9 (\textit{open parcel}) for reference, and the full results and analysis on sub-tasks are provided in \textcolor{black}{Appendix}. 

\noindent\textbf{MLCoT Prompting for Knowledge Acquisition.} To evaluate the impact of core designs in MLCoT, we replace the MLCoT pipeline with the following approaches and utilize the acquired knowledge as the condition to guide detection: 
(1) We encode the object categories returned by the object level of MLCoT as the knowledge to perform the knowledge-conditional object detection. The results (31.9\% mAP$_{\text{box}}$ and 31.3\% mAP$_{\text{mask}}$) demonstrate that simply extracting the object categories from LLMs cannot achieve satisfactory performance. 
(2) We attempt to acquire affordance-level rather than object-level knowledge. Specifically, we prompt LLMs by asking \textit{``what visual features can we use to determine the suitability of an object for \{TASK\}?"} to generate visual affordance knowledge directly. 
The attempt improves the above object-level model by 16.2\% mAP$_{\text{box}}$ and 13.4\% mAP$_{\text{mask}}$, showing the necessity of exploring the essential visual affordances behind the object categories. However, this model still underperforms by approximately 9\% mAP compared to our full model. It is difficult to summarize a unified description of widely varying objects without priors, resulting in only one set of visual attributes being returned from LLMs. 
(3) To increase the diversity of visual affordances, we prompt LLMs to generate visual features for each object retrieved, which leads to a significant improvement to 52.4\% mAP$_{\text{box}}$ and 49.0\% mAP$_{\text{mask}}$. 
(4) Finally, we further add rationales to filter out the misleading and irrelevant attributes, achieving a 4.5\%  and 4.6\% increase in mAP$_{\text{box}}$ and  mAP$_{\text{mask}}$, respectively.
(5) We also evaluate the effect of using different LLMs to extract visual affordance knowledge. Our MLCoT with ChatGPT~\cite{ouyang2022training} has a similar mAP to MLCoT with GPT-3. 

\noindent\textbf{Knowledge-conditional Object Detection.} To validate the effectiveness of our proposed knowledge-conditioned decoder, we conduct ablation studies with two baselines and three variants based on Deformable-DETR~\cite{zhu2020deformable} framework: 
(1) We develop GGNN~\cite{sawatzky2019ggnn} on the Deformable-DETR detection framework. Def+GGNN simply learns the relations between objects and identifies objects based on their contexts, limiting its performance. 
(2) Besides, similar to~TOIST\cite{li2022toist}, we initialize queries with the task's textual feature based on our framework. The performance gap (16.6\% mAP$_{\text{box}}$ and 14.1\% mAP$_{\text{mask}}$) between Def+TOIST and our final model. 
(3) We introduce the visual affordance knowledge extracted by MLCoT but simply use it to initialize the queries of the decoder (Init w/MLCoT). The model achieves significant performance gain compared to the two baselines. 
(4) We further fuse knowledge with the image's visual feature map to construct a multi-modal feature map (Fuse w/MLCoT), which jointly understands the two modalities and improves performance (1.9\% mAP$_{\text{box}}$ and 1.3\% mAP$_{\text{mask}}$) compared to the last model. 
(5) Our proposed knowledge-conditional query generation, generating based on the visual content of the image, the task, and the visual affordance knowledge, helps the decoder better localize the objects, resulting in average improvements of 4.7\% mAP$_{\text{box}}$ and 4.0\% mAP$_{\text{mask}}$.
(6) Finally, the knowledge-conditional denoising training improves AP$_{\text{box}}$ and AP$_{\text{mask}}$ by 1.6\% and 1.9\%, respectively. 
\subsection{Visualization}

Figure~\ref{fig:vz} visualizes qualitative results for several examples. 
For (a), no objects in the image should be selected to \textit{``get lemon out of tea''}.
Our model can successfully return the empty set, while TOIST detects the french fry that is one of the salient objects in the image as the tool. 
Similarly, as knives are uncommon for \textit{``opening bottle of beer''}, the knife in (b) is challenging for TOIST to identify and locate. 
Guided by the visual affordance of \textcolor{black}{``sharp blade with a pointed end"}, our model correctly localizes and selects the sharp knife.
The (c) and (d) demonstrate effectiveness without MLCoT or knowledge-conditional denoising training (KDN). 
With visual affordance knowledge obtained by directly asking LLMs, our model relies solely on matching with the single knowledge unit, which incorrectly detects the trunk in (c) and misses the knife in (d). 
The former trunk is easily confused with objects that are ``flat, broad with a handle", while the latter knife is ignored because its visual attributes of straight mismatch the single knowledge unit that includes ``curved or angled". 
Furthermore, without KDN, our detector lacks explicit guidance, leading to inaccurate detection in challenging scenes.
Specifically, the glove in (c) and the knife in (d) are not detected successfully, and the packing line in (d) is mistakenly detected.
\section{Conclusion}
In this paper, we focus on challenging task driven object detection, which is practical in the real world yet under-explored. To bridge the gap between abstract task requirements and objects in the image, we propose to explicitly extract visual affordance knowledge for the task and detect objects having consistent visual attributes to the visual knowledge. Furthermore, our CoTDet utilizes visual affordance knowledge to condition the decoder in localizing and recognizing suitable objects.

\noindent\textbf{Limitations:} While acknowledging the disparity between the COCO-Task dataset and real-world application scenarios, attributed to its limited task variety and preference for images and annotations, our approach has the potential to extend beyond these confines. 
Notably, our knowledge acquisition and utilization are flexible and generalizable, granting it the capacity to transcend specific dataset, specific tasks, object categories, or tools. We leave this to future works. 
Furthermore, with the incorporation of LLM, our approach inherits potential social biases from LLM, which could potentially be reflected in the preference for selecting frequently used tools.

\noindent\textbf{Acknowledgment:} This work was supported by the National Natural Science Foundation of China (No.62206174), Shanghai Pujiang Program (No.21PJ1410900), Shanghai Frontiers Science Center of Human-centered Artificial Intelligence (ShangHAI), MoE Key Laboratory of Intelligent Perception and Human-Machine Collaboration (ShanghaiTech University), and Shanghai Engineering Research Center of Intelligent Vision and Imaging.


{\small
\bibliographystyle{ieee_fullname}
\bibliography{egbib}
}

\end{document}